# Unstructuring User Preferences: Efficient Non-Parametric Utility Revelation

**Carmel Domshlak**
Fac. of Industrial Engineering & Management
Technion - Israel Institute of Technology
Haifa, Israel 32000

**Thorsten Joachims**
Computer Science Dept.
Cornell University
Ithaca, NY 14853

## Abstract

Tackling the problem of ordinal preference revelation and reasoning, we propose a novel methodology for generating an ordinal utility function from a set of qualitative preference statements. To the best of our knowledge, our proposal constitutes the first non-parametric solution for this problem that is both efficient and semantically sound. Our initial experiments provide strong evidence for practical effectiveness of our approach.

## 1 INTRODUCTION

Human preferences are a key concept in decision theory, and as such have been studied extensively in philosophy, psychology, and economics (e.g., [10, 12, 14]). The central goals have been to provide logical, cognitive, and mathematical models of human decision making. More recently this research effort was joined by AI researchers, motivated by the goal of *automating* the process of decision support. To illustrate the need for automated decision support, consider the nowadays common task of searching for some goods in a database of an online vendor such as Amazon.com or eBay. Such databases are too large for a user to search exhaustively. Using the purchase of a used car as an example, a decision support system might allow a user to state preferences like "I like ecologically friendly cars", "I prefer Mercedes to Lada", or "For a sport car, I prefer red color to black color". The system should then use these preference statements to guide the user to the relevant parts of the database.

Various logics of preference, graphical preference representation models, preference learning and reasoning algorithms have been proposed in AI in the last three decades [4, 8, 9]. While these works have made significant contributions, there is still a substantial gap between theory and practice of decision support. In particular, so far there is no single framework for revealing user preferences and reasoning about them that is both generically scalable and generically robust, i.e. *both* efficient and effective for *any* set of decision alternatives and *any* form of preference information. It is clear nowadays that getting closer to such a universal framework requires new insights into the problem [8, 21].

In this paper, we tackle this challenge in the scope of revelation and reasoning about *ordinal preferences* (i.e., as in the database search example above), and develop a robust solution for this problem that is both efficient and effective. Specifically, we propose a novel methodology for generating an ordinal utility function from a set of qualitative preference statements. Our proposal is based on a somewhat surprising mixture of techniques from knowledge representation and machine learning. We show formally that it leads to a flexible and unprecedentedly powerful tool for reasoning about ordinal preference statements. Furthermore, we present experiments that provide initial evidence for practical applicability and effectiveness of our method, making it promising for a wide spectrum of decision-support applications.

### 1.1 PROBLEM STATEMENT AND BACKGROUND

Using the used-car database search as our running example, the content of the database constitutes the relevant subset of all possible choice alternatives $\Omega$. The ordinal preferences of a user who wants to buy a car can be viewed as a (possibly weak, possibly partial) binary preference relation $P$ over $\Omega$ [12]. A decision support system should allow its user to state her preferences, use these statements to approximate $P$, and present the database content in a way that enables the user to quickly home in on desirable alternatives.

The choice alternatives in such scenarios are typically described in terms of some *attribution* $\mathbf{X} = \{X_1, \ldots, X_n\}$ abstracting $\Omega$ to $\mathcal{X} = \times Dom(X_i)$ (e.g., attributes of the database schema), and the user can express her preferences in terms of $\mathbf{X}$. Now, what preference information can we expect the users to provide? As suggested in the used-car example (and supported by multi-disciplinary literature [8, 12]), typically users should be expected to provide only qualitative preference statements that either compare between pairs of complete alternatives (e.g., "I prefer this alternative to that alternative"), or generalize user's preference over some properties of $\Omega$ (e.g., "In a minivan, I prefer automatic transmission to manual transmission.") Formally, this means that the user provides us with a qualitative preference expression

$$S = \{s_1, \ldots, s_m\} = \{\langle \varphi_1 \oslash_1 \psi_1 \rangle, \cdots, \langle \varphi_m \oslash_m \psi_m \rangle\}, \quad (1)$$

consisting of a set of such preference statements $\{s_1, \ldots, s_m\}$, where $\varphi_i, \psi_i$ are logical formulas, $\oslash_i \in \{\succ, \succeq, \sim\}$, and $\succ, \succeq$, and $\sim$ have the standard semantics of strong preference, weak preference, and preferential equivalence, respectively. For ease of presentation, we assume attributes $\mathbf{X}$ are boolean[1] (denoting $Dom(X_i) = \{x_i, \overline{x_i}\}$), and $\varphi_i, \psi_i$ are propositional logic formulas over $\mathbf{X}$.

Given such a preference expression $S$, one has to interpret what information $S$ conveys about $P$, decide on a representation for this information, and decide on the actual reasoning machinery. Several proposals for direct logical reasoning about $S$ have been made, yet all these proposals are limited by (this or another) efficiency/expressiveness tradeoff. In attempt to escape this tradeoff as much as possible, several works in AI (e.g., see [5, 18]) proposed to compile information carried by $S$ into an ordinal utility function

$$U : \mathcal{X} \mapsto \mathbb{R} \quad (2)$$

consistent with (what we believe $S$ tells us about) $P$, that is requiring

$$\forall \mathbf{x}, \mathbf{x}' \in \mathcal{X}. \ U(\mathbf{x}) \geq U(\mathbf{x}') \Rightarrow P \not\models \mathbf{x}' \succ \mathbf{x}. \quad (3)$$

In what follows, we refer to the task of constructing such a utility function $U$ from $S$ as *ordinal utility revelation (OUR)*. Observe that specifying a utility function $U$ as in (2) can be expensive due to the fact that $|\mathcal{X}| = O(2^n)$. Hence, previous works on OUR searched for special conditions under which $U$ can be represented compactly (e.g., see [1, 3, 5, 11, 17, 18]). The general scheme followed by these works (which we refer to as independence-based methodology) is as follows. First, one defines certain "independence conditions" on $\mathbf{X}$, and provides a "representation theorem" stating that under these conditions $U$ can be compactly specified. Second, one possibly defines some additional conditions under which $U$ can also be efficiently generated from $S$. Finally, assuming all these conditions lead to restricting user expressions $S$ to simplified languages that incorporate this prior assumption. In summary, computationally efficient schemes for multi-attribute utility revelation proposed in economics and AI are parametrized by the *structure* that user preferences induce on $\mathbf{X}$, and thus are applicable only when such structure *exists* and is *known* to the system.

---
[1]Extending our framework to arbitrary finite-domain variables is straightforward, yet requires a more involved notation that we decided to avoid here.

## 1.2 CHALLENGES AND OUR RESULTS

Having in mind these limitations, let us return back to the needs of decision-support application, and list the challenges these applications pose to the research on OUR. The vision here is threefold. First, the user should be able to provide preference expressions $S$ while being as little constrained in her language as possible. Second, the utility revelation machinery should be completely non-parametric, i.e., free of any explicit assumptions about the structure of user preferences. Third, both utility revelation (i.e., generating $U$ from $S$) and using the revealed utility function should be computationally efficient, including the case where user preferences pose no significant independence structure on $\mathbf{X}$ whatsoever.

In this paper, we present the first approach that fulfills these goals. Combining ideas from knowledge representation, machine learning, and philosophical logic we provide a concrete mathematical setting in which all the above desiderata can be successfully achieved, and formally show that this setting is appealing both semantically and computationally. The mathematical framework we propose is based on a novel high-dimensional structure for preference decomposition, and a specific adaptation of certain standard techniques for high-dimensional continuous optimization, frequently used in machine learning in the context of Support Vector Machines (SVMs) [22].

## 2 HIGH-DIMENSIONAL PREFERENCE DECOMPOSITION

Considering our vision for preference revelation, one can certainly be somewhat skeptical. Indeed, how can OUR be efficient if the user preferences pose no significant independence structure on $\mathbf{X}$, or, if they do, the system is not provided with this independence information? The basic idea underlying our proposal is simple: *Since we are not provided with a sufficiently useful independence information in the original space $\mathcal{X}$, maybe we should move to a different space in which no independence information is required?*

Specifically, let us schematically map the alternatives

$\mathcal{X}$ into a new, higher dimensional space $\mathcal{F}$ using

$$\Phi : \mathcal{X} \mapsto \mathcal{F} = \mathbb{R}^{4^n} . \qquad (4)$$

As one would expect, the mapping $\Phi$ is not arbitrary. Let $\mathbf{F} = \{\mathfrak{f}_1, \cdots, \mathfrak{f}_{4^n}\}$ be the dimensions of $\mathcal{F}$, and $D = \bigcup Dom(X_i)$ be the union of attribute domains in $\mathbf{X}$. Let $\mathsf{val} : \mathbf{F} \to 2^D$ be a bijective mapping from the dimensions of $\mathcal{F}$ onto the power set of $D$, uniquely associating each dimension $\mathfrak{f}_i$ with a subset $\mathsf{val}(\mathfrak{f}_i) \subseteq \{x_1, \overline{x_1}, \cdots, x_n, \overline{x_n}\}$. In what follows, by $\mathsf{Var}(\mathfrak{f}_i) \subseteq \mathbf{X}$ we denote the subset of attributes "instantiated" by $\mathsf{val}(\mathfrak{f}_i)$. Given that, for each $\mathbf{x} \in \mathcal{X}$ and $\mathfrak{f}_i \in \mathbf{F}$, we set:

$$\Phi(\mathbf{x})[i] = \begin{cases} 1, & \mathsf{val}(\mathfrak{f}_i) \subseteq \mathbf{x} \\ 0, & \text{otherwise} \end{cases} \qquad (5)$$

From (5) it is easy to see that dimensions $\mathfrak{f}_i$ with $\mathsf{val}(\mathfrak{f}_i)$ containing both a literal and its negation are effectively redundant. Indeed, later we show that we actually use only the $(3^n - 1)$-dimensional subspace of $\mathcal{F}$, dimensions of which correspond to all the *non-empty partial assignments on* $\mathbf{X}$. Hence, for ease of presentation, in what follows we discuss $\mathcal{F}$ as if ignoring its redundant dimensions. However, for some technical reasons important for our computational machinery, the structure of $\mathcal{F}$ and $\Phi$ has to be defined as in (4)-(5).

To illustrate our mapping $\Phi$, if $\mathbf{X} = \{X_1, X_2\}$ and $\mathbf{x} = x_1 \overline{x_2}$, we have $\Phi(\mathbf{x})[i] = 1$ if and only if $\mathsf{val}(\mathfrak{f}_i) \in \{x_1, \overline{x_2}, x_1 \overline{x_2}\}$, that is

$$\Phi(\mathbf{x}) = \begin{pmatrix} 1 \\ 0 \\ 0 \\ 1 \\ 0 \\ 1 \\ 0 \\ 0 \end{pmatrix} \begin{matrix} \mathsf{val}(\mathfrak{f}_1) = x_1 \\ \mathsf{val}(\mathfrak{f}_2) = \overline{x_1} \\ \mathsf{val}(\mathfrak{f}_3) = x_2 \\ \mathsf{val}(\mathfrak{f}_4) = \overline{x_2} \\ \mathsf{val}(\mathfrak{f}_5) = x_1 x_2 \\ \mathsf{val}(\mathfrak{f}_6) = x_1 \overline{x_2} \\ \mathsf{val}(\mathfrak{f}_7) = \overline{x_1} x_2 \\ \mathsf{val}(\mathfrak{f}_8) = \overline{x_1} \overline{x_2} \end{matrix} \qquad (6)$$

where (6) addresses only the non-redundant dimensions of $\mathcal{F}$.

Geometrically, $\Phi$ maps each $n$-dimensional vector $\mathbf{x} \in \mathcal{X}$ to the $4^n$-dimensional vector in $\mathcal{F}$ that uniquely encodes the set of all projections of $\mathbf{x}$ onto the subspaces of $\mathcal{X}$. But is $\mathcal{F}$ semantically intuitive? After all, why should we adopt this and not some another dimensional structure for $\Omega$? To answer this question, recall that $\mathbf{X}$ is just an attribution of $\Omega$ (induced by some application-dependent considerations), and as such it does not necessarily correspond to the criteria affecting preference of the user over the actual physical alternatives. However, if the user provides us with some preference statements in terms of $\mathbf{X}$, the implicit criteria behind these statements obviously have some encoding in terms of $\mathbf{X}$. Given that, the semantic attractiveness of $\mathcal{F}$ is apparent: it is not hard to see that evaluation of *any* such implicit, preference-related criterion on $\mathbf{x} \in \mathcal{X}$ necessarily corresponds to a single dimension of $\mathcal{F}$. In addition, Theorem 1 shows that $\mathcal{F}$ is not only semantically intuitive, but also satisfies our requirement of "no need for independence information".

**Theorem 1** *Any preference ordering $P$ over $\mathcal{X}$ is additively decomposable in $\mathcal{F}$, that is, the existence of a linear function*

$$\mathfrak{U}(\Phi(\mathbf{x})) = \sum_{i=1}^{4^n} w_i \; \Phi(\mathbf{x})[i] \qquad (7)$$

*satisfying (3) is guaranteed for any such $P$ over $\mathcal{X}$.*

The proof of Theorem 1 is straightforward since we can always specify weights $w_i$ associated with all complete assignments to $\mathbf{X}$ such that (3) is satisfied. It is important to note, however, that this explicit "construction" of $\mathfrak{U}$ only serves the *existential* proof of Theorem 1, and does not reflect whatsoever the machinery of our proposal.

Since, by Theorem 1, dimensions $\mathcal{F}$ can successfully "linearize" any preference ordering $P$ over $\mathcal{X}$, in what follows we can focus only on linear utility functions as in (7). Of course, the reader may rightfully wonder whether this linearization in a space of dimension $4^n$ can be of any practical use, and not just a syntactic sugar. However, at this stage we ask the reader to postpone the computational concerns, and focus on the *interpretation* of preference expressions in the scope of our new high dimensional space $\mathcal{F}$.

There are two major categories of preference statements one would certainly like to allow in $S$ [12], notably dyadic (comparative) statements (indicating a relation between two referents using the concepts such as 'better', 'worse', and 'equal in value to'), and monadic (classificatory) statements (evaluating a single referent using ordinal language concepts such as 'good', 'very bad', and 'worst'.)[2] For ease of presentation, let us focus on dyadic statements for now. In particular, consider an "instance comparison" statement "$\mathbf{x}$ is better than $\mathbf{x}'$", where $\mathbf{x}, \mathbf{x}' \in \mathcal{X}$. The interpretation of this statement poses no serious difficulties because it explicitly compares between complete descriptions of two alternatives. However, this is the exception, rather than the rule. Most of the preference statements that we use in our everyday activities (e.g., "I prefer compact cars to SUVs") have this or another generalizing nature. As such, these statements typically mention only a subset of attributes. This creates an ambiguity with respect to their actual referents. Several proposals on how to interpret preference statements have been made both in philosophy and AI, but there is no (and cannot be?) an agreed-upon solution to this problem [12]. However, all the proposals suggest to interpret generalizing preference statements as comparing between complete descriptions $\mathcal{X}$ of the alternatives, while disagreeing on what complete descriptions are actually compared by each statement separately, and/or by a multi-statement preference expression as a whole.

---

[2]This classification does not cover more "higher order" preferences, such as "$x$ is preferred to $y$ more than $z$ is preferred to $w$" [19]. Although here we do not discuss such statements, they as well can be processed in our framework.

Considering interpretation of qualitative preference expressions in $\mathcal{F}$, observe that each parameter $w_i$ of $\mathfrak{U}$ as in (7) can be seen as capturing the *marginal value of the interaction* between $\mathsf{Var}(\mathfrak{f}_i)$ when these take the value $\mathsf{val}(\mathfrak{f}_i)$. Note that $w_i$ corresponds to this specific interaction *only*; all the syntactically related interactions of subsets and supersets of $\mathsf{val}(\mathfrak{f}_i)$ are captured by other parameters $w$, and the dimensional structure of $\mathcal{F}$ allows such an independent bookkeeping of all possible value-related criteria.

Now, consider an arbitrary dyadic statement $\varphi \succ \psi$. Let $\mathbf{X}_\varphi \subseteq \mathbf{X}$ (and similarly $\mathbf{X}_\psi$) be the variables involved in $\varphi$, and $M(\varphi) \subseteq Dom(\mathbf{X}_\varphi)$ be the set of $\varphi$'s models. Following the most standard (if not the only) interpretation scheme for OUR, we compile $\varphi \succ \psi$ into a set of constraints on the space of candidate utility functions [15]. In our case, however, these constraints are posed on the functions of form (7), which are linear, real-valued functions from the feature space $\mathcal{F}$, and not from the original attribute space $\mathcal{X}$ as in previous works. Specifically, we compile $\varphi \succ \psi$ into a set of $|M(\varphi)| \times |M(\psi)|$ constraints [3]

$$\forall m \in M(\varphi), \forall m' \in M(\psi). \sum_{\mathfrak{f}_i:\mathsf{val}(\mathfrak{f}_i) \in 2^m} w_i > \sum_{\mathfrak{f}_j:\mathsf{val}(\mathfrak{f}_j) \in 2^{m'}} w_j \quad (8)$$

where $2^m$ denotes the set of all value subsets of $m$. For example, statement $(X_1 \vee X_2) \succ (\neg X_3)$ (e.g., "It is more important that the car is powerful or fast than not having had an accident") is compiled into

$$w_{x_1} + w_{x_2} + w_{x_1 x_2} > w_{\overline{x_3}}$$
$$w_{x_1} + w_{\overline{x_2}} + w_{x_1 \overline{x_2}} > w_{\overline{x_3}} \quad (9)$$
$$w_{\overline{x_1}} + w_{x_2} + w_{\overline{x_1} x_2} > w_{\overline{x_3}}$$

The constraint system $\mathfrak{C}$ resulting from such compilation of a user expression $S$ defines the space of solutions for our formulation of OUR. On the side of the semantics, we argue that $\mathfrak{C}$ corresponds to a *least committing interpretation* of preference statements. This encodes the principle that, if there is no reason for a bias towards certain explanations for $\varphi \succ \psi$, a most general explanation should be preferred. In Section 3 we describe how we pick a particular assignment to $w_i$ for a given set of constraints, and justify this choice in Section 3.1.

In first view, we clearly have some complexity issues here. First, while the constraint system $\mathfrak{C}$ is linear, it is linear in the exponential space $\mathbb{R}^{4^n}$. Second, the summations in *each* constraint as in (8) are exponential in the arity of $\varphi$ and $\psi$ (i.e., in $|\mathbf{X}_\varphi|$ and $|\mathbf{X}_\psi|$). Finally, the number of constraints generated for each preference statement can be exponential in the arity of $\varphi$ and $\psi$ as well.

While exponential dimensionality of $\mathcal{F}$ is inherit in our framework (and we promised to do something about it later), the description complexity of $\mathfrak{C}$ deserves a closer look. First, the description size of each constraint is clearly something to worry about. For instance, each "instance comparison" between a pair of complete alternatives in $\mathcal{X}$ is translated into a constraint with up to $2^{n+1}$ summation terms, and this is a very natural form of everyday preference statements. Fortunately, in Section 3 we efficiently overcome this obstacle. On the other hand, the number of constraints per preference statement seems to be less problematic in practice, because the number of constraints equals the number of models of $\varphi$ and $\psi$, and explicit simultaneous preferential comparison between large sets of models are rarely natural.

## 3 COMPUTATIONAL MACHINERY

At this point, we hope to have convinced the reader that semantically our construction is appealing. What still remains to be shown is that it is computationally realistic. We begin with summarizing the complexity issues that we have to resolve.

(a) Our target utility function $\mathfrak{U}$ is a linear, real-valued function from a $4^n$ dimensional space $\mathcal{F}$. Thus, not only generating $\mathfrak{U}$, but even keeping and evaluating this function explicitly might be infeasible.

(b) The space of all suitable functions $\mathfrak{U}$ is defined by a set of linear constraints $\mathfrak{C}$ in $\mathbb{R}^{4^n}$. In addition to the dimensionality of this satisfiability problem, even the description of each constraint can be exponential in $n = |\mathbf{X}|$ for many natural preference statements.

In the following we show that both these complexity issues can be overcome. For ease of presentation and without loss of generality, we introduce our machinery on preference expressions consisting only of strict

---
[3]The constraints for dyadic statements of the form $\varphi \succeq \psi$ and $\varphi \sim \psi$ are similar to (8) with $>$ being replaced by $\geq$ and $=$, respectively.

"instance comparisons" $\mathbf{x} \succ \mathbf{x}'$, where $\mathbf{x}, \mathbf{x}' \in \mathbf{X}$. Our translation of each such dyadic preference statement $\mathbf{x} \succ \mathbf{x}'$ leads to a linear constraint of the form:

$$\mathfrak{U}(\Phi(\mathbf{x})) > \mathfrak{U}(\Phi(\mathbf{x}')) \Leftrightarrow \sum_{i=1}^{4^n} w_i \, \Phi(\mathbf{x})[i] > \sum_{i=1}^{4^n} w_i \, \Phi(\mathbf{x}')[i]$$
$$\Leftrightarrow \mathbf{w} \cdot \Phi(\mathbf{x}) > \mathbf{w} \cdot \Phi(\mathbf{x}') \quad (10)$$

According to this formulation, the set of utility functions consistent with a set of $k$ such preference statements is defined by the solutions of the linear system $\mathfrak{C}$:

$$\forall 1 \leq i \leq k. \ \mathbf{w} \cdot \Phi(\mathbf{x}_i) > \mathbf{w} \cdot \Phi(\mathbf{x}'_i), \quad (11)$$

consisting of $k$ constraints in $\mathbb{R}^{4^n}$. Clearly, naive approaches to solving such systems will be computationally intractable for interesting $n$. In what follows, we will exploit duality techniques from optimization theory (see [2]) and Mercer kernels (see [22]) as used in machine learning to solve such systems in time that is linear in $n$ and polynomial in $k$.

At the first step, we reformulate our task of satisfying $\mathfrak{C}$ as an optimization problem. Since the solution of (11) is typically not unique, we select a particular solution by adding an objective function and a "margin" by which the inequality constraints should be fulfilled. Specifically, similar to an ordinal regression SVM [13], we search for the smallest $L_2$ weight vector $\mathbf{w}$ that fulfills all constraints with margin 1. The corresponding constrained optimization problem is:

$$\text{Minimize } (w.r.t. \ \mathbf{w}) : \frac{1}{2} \mathbf{w} \cdot \mathbf{w}$$
$$\text{subject to :} \quad (12)$$
$$\forall 1 \leq i \leq k. \ \mathbf{w} \cdot \Phi(\mathbf{x}_i) \geq \mathbf{w} \cdot \Phi(\mathbf{x}'_i) + 1$$

Note that this reformulation of the problem does not affect its satisfiability, and that the solution of (12) is unique, since it is a strictly convex quadratic program.

In the second step we consider the Wolfe dual [2] of (12):

$$\text{Maximize } (w.r.t. \ \alpha) : \quad (13)$$
$$\sum_{i=1}^{k} \alpha_i - \frac{1}{2} \sum_{i=1}^{k} \sum_{j=1}^{k} \alpha_i \alpha_j ((\Phi(\mathbf{x}_i) - \Phi(\mathbf{x}'_i)) \cdot (\Phi(\mathbf{x}_j) - \Phi(\mathbf{x}'_j)))$$
$$\text{subject to : } \alpha \geq \mathbf{0}$$

This is a standard technique frequently used in the context of SVMs [22, 13]. The Wolfe dual (13) has the same optimum value as the primal (12). From the parameter vector $\alpha^*$ that solves the dual one can derive the solution $\mathbf{w}^*$ of the primal as $\mathbf{w}^* = \sum_{i=1}^{m} \alpha_i^* (\Phi(\mathbf{x}_i) - \Phi(\mathbf{x}'_i))$.

The third and final step is based on the observation that the dual (13) can be expressed in terms of inner products in the high-dimensional feature space. For many kinds of mappings $\Phi$, inner products can be computed efficiently using a Mercer kernel (see [22]), even if $\Phi$ maps into a high-dimensional (or infinite dimensional) space. Our task, thus, is to find such a kernel for the specific mapping $\Phi$ that we use in our construction (4)-(5).

Let us define an injective representation of attribute vectors $\mathbf{x}$ by projecting them to indicator vectors $\vec{\mathbb{x}} \in \mathbb{R}^{2n}$. Each attribute value is mapped onto a single dimension. If an attribute value is present in $\mathbf{x}$, the corresponding component of $\vec{\mathbb{x}}$ is 1, otherwise 0. If an attribute is unspecified, all corresponding components of $\vec{\mathbb{x}}$ are set to 0. Using this construction, inner products for an (effectively equivalent) variant $\Phi_\lambda$ of our mapping $\Phi$ can be computed as follows.

**Theorem 2** *For the mapping* $\Phi_\lambda : \mathcal{X} \mapsto \mathcal{F} = \mathbb{R}^{4^n}$

$$\Phi_\lambda(\mathbf{x})[i] = \begin{cases} \sqrt{c_\lambda(|\mathsf{val}(\mathfrak{f}_i)|)}, & \mathsf{val}(\mathfrak{f}_i) \subseteq \mathbf{x} \\ 0, & \text{otherwise} \end{cases} \quad (14)$$

*where*

$$c_\lambda(k) = \sum_{l=k}^{n} \lambda_l \sum_{\substack{l_1 \geq 1, \ldots, l_k \geq 1 \\ l_1 + \ldots + l_k = l}} \frac{l!}{l_1! \ldots l_k!}, \quad (15)$$

*and any* $\mathbf{x}, \mathbf{x}' \in \mathcal{X}$ *and* $\lambda_1, \ldots, \lambda_n \geq 0$, *the kernel*

$$K(\mathbf{x}, \mathbf{x}') = \sum_{l=1}^{n} \lambda_l (\vec{\mathbb{x}} \cdot \vec{\mathbb{x}}')^l \quad (16)$$

*computes the inner product* $\Phi_\lambda(\mathbf{x}) \cdot \Phi_\lambda(\mathbf{x}') = K(\mathbf{x}, \mathbf{x}')$.

**Proof** *The following chain of equalities holds.*

$$K(\mathbf{x}, \mathbf{x}') = \sum_{l=1}^{n} \lambda_l (\mathbb{x} \cdot \mathbb{x}')^l$$
$$= \sum_{l=1}^{n} \lambda_l \sum_{(i_1, \ldots, i_l) \in \{1, \ldots, 2n\}^l} (\mathbb{x}_{i_1} \mathbb{x}'_{i_1} \mathbb{x}_{i_2} \mathbb{x}'_{i_2} \ldots \mathbb{x}_{i_l} \mathbb{x}'_{i_l})$$
$$= \sum_{l=1}^{n} \lambda_l \sum_{(i_1, \ldots, i_l) \in \{1, \ldots, 2n\}^l} (\mathbb{x}_{i_1} \mathbb{x}_{i_2} \ldots \mathbb{x}_{i_l})(\mathbb{x}'_{i_1} \mathbb{x}'_{i_2} \ldots \mathbb{x}'_{i_l})$$
$$= \sum_{k=1}^{n} c_\lambda(k) \sum_{\{i_1, \ldots, i_k\} \subseteq \{1, \ldots, 2n\}} (\mathbb{x}_{i_1} \mathbb{x}_{i_2} \ldots \mathbb{x}_{i_k})(\mathbb{x}'_{i_1} \mathbb{x}'_{i_2} \ldots \mathbb{x}'_{i_k})$$
$$= \Phi_\lambda(\mathbf{x}) \cdot \Phi_\lambda(\mathbf{x}')$$

*$c_\lambda(k)$ is the multiplicity with which a monomial of size $k$ occurs. The multiplicity is influenced by two factors. First, different orderings of the index sequence $(i_1, \ldots, i_l)$ lead to the same term. This is counted by the multinomial coefficient $\frac{l!}{l_1! \ldots l_k!}$, where $l_1, \ldots, l_k$ are the powers of each factor. Second, all positive powers of any $\mathbb{x}_i \mathbb{x}'_i$ are equal. We therefore sum over all such*

*equivalent terms*

$$\sum_{\substack{l_1 \geq 1, \dots, l_k \geq 1 \\ l_1 + \dots + l_k = l}} \frac{l!}{l_1! \dots l_k!}$$

*Note that many of the monomials always evaluate to zero under our encoding $\vec{x}$ of $\mathbf{x}$. Specifically, monomials corresponding to expressions $(x_i \wedge \overline{x_i} \wedge \cdots)$ will always be nullified. In particular, it is therefore sufficient to consider only those monomials of size less or equal to $n$, since all others will always evaluate to zero.* ∎

The kernel (16), which is similar to a polynomial kernel [22], allows us to compute inner products in the high-dimensional space in linear time, and, for strictly positive $\lambda_1, \dots, \lambda_n$, moving from $\Phi$ to $\Phi_\lambda$ does not change the satisfiability of our constraint system $\mathfrak{C}$. To see the latter, observe that any solution $\mathbf{w}_\lambda$ of (11) for $\Phi_\lambda$ corresponds to a solution $\mathbf{w}$ of (11) for $\Phi$ via

$$\mathbf{w}_\lambda[i] = \frac{\mathbf{w}[i]}{\sqrt{c_\lambda(|\mathsf{val}(\mathfrak{f}_i)|)}}.$$

The difference between the mappings $\Phi_\lambda$ and $\Phi$ is that the former biases the inference's prior towards smaller size monomials, "preferring" more general explanations for user preference statements. On the other hand, this bias can be controlled to a large degree via the kernel parameters $\lambda_1, \dots, \lambda_n$.

Now, using the kernel inside of the dual leads to the following equivalent optimization problem.

*Maximize (w.r.t. $\alpha$)* :

$$\sum_{i=1}^{k} \alpha_i - \frac{1}{2} \sum_{i=1}^{k} \sum_{j=1}^{k} \alpha_i \alpha_j (K(\mathbf{x}_i, \mathbf{x}_j) - K(\mathbf{x}_i, \mathbf{x}'_j) - K(\mathbf{x}'_i, \mathbf{x}_j) + K(\mathbf{x}'_i, \mathbf{x}'_j)) \quad (17)$$

*subject to* : $\alpha \geq \mathbf{0}$

It is known that such convex quadratic programs can be solved in polynomial time [2]. To compute the value of $\mathfrak{U}$ for a given alternative $\mathbf{x}'' \in \mathcal{X}$, it is sufficient to know only the dual solution and the kernel:

$$\mathfrak{U}(\Phi(\mathbf{x}'')) = \mathbf{w}^* \cdot \Phi_\lambda(\mathbf{x}'') = \sum_{i=1}^{k} \alpha_i (K(\mathbf{x}_i, \mathbf{x}'') - K(\mathbf{x}'_i, \mathbf{x}''))$$
(18)

Hence, *neither computing the solutions of the constraint system $\mathfrak{C}$, nor computing the values of $\mathfrak{U}$ on $\mathcal{X}$ requires any explicit computations in $\mathbb{R}^{4^n}$*. Through the use of kernels, all computations can be done efficiently in the low-dimensional input space.[4]

---

[4] We have extended $SVM^{light}$ to solve this type of quadratic optimization problem. The implementation is available at http://svmlight.joachims.org/. It can efficiently handle large-scale problems with $n, m \approx 10,000$.

As a final comment on the mechanics of our inference procedure, note that it would be unreasonable to expect that a user's preference statements will always be consistent. In case of inconsistent preference specification, one can use the standard soft-margin technique [7], trading-off constraint violations against margin size.

### 3.1 INFERENCE SEMANTICS

Since the user's statements typically provide only partial information about her preferences, the constraint system in (11) is underconstrained, and thus the utility revelation takes the form of inductive reasoning. If the system has access to a prior $Pr(\mathfrak{U})$ over utility functions, a reasonable inductive inference procedure would be to pick the most likely utility function $\mathfrak{U}$ that fulfills all constraints. In particular, for the Gaussian prior $Pr(\mathfrak{U}) \sim e^{||\mathbf{w}||^2}$ this procedure results in finding the weight vector with minimum $L_2$-norm that fulfills the constraints. This is exactly our objective in (12). To illustrate the behavior arising from this prior, consider the statements

$$\begin{aligned} s_1 &= (X_1 \vee X_2) \succ (\neg X_3), \\ s_2 &= (X_3) \succ (X_4), \\ s_3 &= (X_1) \succ (X_2). \end{aligned}$$

For this small set of constraints, we can compute the solution without the use of kernel and get the following weights.

$$\begin{array}{llll} w_{x_1}{=}0.75 & w_{x_2}{=}{-}0.25 & w_{x_3}{=}0.5 & w_{x_4}{=}{-}0.5 \\ w_{\overline{x_1}}{=}0.4 & w_{\overline{x_2}}{=}0 & w_{\overline{x_3}}{=}{-}0.45 & w_{\overline{x_4}}{=}0 \\ w_{x_1 x_2}{=}0.05 & w_{\overline{x_1} x_2}{=}0.4 & & \end{array}$$

All other weights are set to zero. Below is an illustrative excerpt of the ordering induced by the utility function generated in our framework:

$$\begin{aligned} \mathfrak{U}(\Phi(x_1 \overline{x_2} x_3 \overline{x_4})) &= 1.25 \\ \mathfrak{U}(\Phi(x_1 x_2 x_3 \overline{x_4})) &= 1.05 \\ \mathfrak{U}(\Phi(\overline{x_1 x_2} x_3 \overline{x_4})) &= 0.9 \\ \mathfrak{U}(\Phi(x_1 x_2 x_3 x_4)) &= 0.55 \\ \mathfrak{U}(\Phi(x_1 x_2 \overline{x_3 x_4})) &= 0.1 \\ \mathfrak{U}(\Phi(x_1 x_2 \overline{x_3} x_4)) &= -0.4 \\ \mathfrak{U}(\Phi(\overline{x_1 x_2 x_3} x_4)) &= -0.55 \end{aligned}$$

We believe that this ordering reflects a natural interpretation of the statements. Furthermore, alternatives for which the statements give no clear judgment receive utility values closer to zero than those for which a statement clearly applies. In general, the Gaussian prior appears reasonable in situations where we expect the utility function to have a compact form (i.e. the weights in $\mathbf{w}$ are small).

Now, recall that (i) each $w_i$ is devoted to capture the marginal value of the event $\mathsf{val}(\mathfrak{f}_i)$, and that (ii) we strive to a least committing interpretation of preference expressions. Observe that our inference procedure implicitly provides us with a reference point $\mathbf{0} \in \mathbb{R}^{4^n}$. In short, we have $w_i = 0$ in case there is no reason to believe the user associates some (positive/negative) value with $\mathsf{val}(\mathfrak{f}_i)$. Thus, consistent with standard logics of monadic preference concepts [6], utility $\mathfrak{U}(\Phi(\mathbf{x})) = 0$ indicates that a user is either indifferent about $\mathbf{x}$ (i.e., has no reason to like it or dislike it), or neutral about it (user's reasons to like $\mathbf{x}$ somehow "balance" her reasons to dislike it). Moreover, this reference point provides us with an intuitive encoding of monadic preference statements. For instance, a statement "$\varphi$ is good" is translated into a set of $|M(\varphi)|$ constraints $\sum_{\mathfrak{f}_i:\mathsf{val}(\mathfrak{f}_i)\in 2^m} w_i > 0$, which can be seen as a special case of (8).

## 4 EVALUATION OF EMPIRICAL EFFECTIVENESS

To demonstrate practical effectiveness of our approach, we conducted experiments on the EachMovie dataset[5]. The dataset consists of six-point-scale movie ratings collected from 72916 users on a corpus of 1628 movies. Each movie is described by a set of attributes, out of which we use the decade of the movie, whether it is currently in the movie theaters, and a binary classification according to ten (non-disjoint) genre categories. In our experiments we generate one ordinal utility function for each user.

The EachMovie dataset contains ratings for individual movies, but no generalizing preference statements. To simulate generalizing preference statements, we generated such statements using the C4.5 decision trees learning algorithm [20] on the following binary classification problem. As training examples, we form all pairs of movies by concatenating their attribute vectors. For each user we generate a separate training set. If the first movie was rated higher (lower) than the second movie, the pair is labeled positive (negative). No pair is generated if at least one of the movies was not rated or if both movies have the same rating, since it was unclear how to translate such cases into training examples for the classification task. On this data, we run the C4.5 decision tree learner[6]. Using the `c45rules` software included in the C4.5 package we then convert the resulting decision tree into a set of rules ordered by their level of confidence, and interpret each of these learned rules as a single preference statement. For example, the highest ranked rule for the user that rated the largest number of movies was the rule (a) below.

(a)
```
B_decade = 90s
B_Art_Foreign = 1
B_Family = 0
B_Romance = 0
  -> A preferred to B
```

(b)
```
A_decade = 80s
A_Thriller = 1
B_Classic = 0
B_Horror = 1
  -> A preferred to B
```

This rule can be interpreted as the monadic preference statement "the user does not like foreign films from the 90s that are not Romance or Family movies". For the same user, the highest ranked dyadic rule is rule (b), meaning "the user prefers thrillers from the 80s over non-classic horror movies".

The quality of the orderings induced by the generated utility functions is measured in terms of ordering error, that is the fraction of times where the user rating and the utility function disagree on the ordering of two movies. For this error measure we consider only movie pairs unequally rated by the user. Ties in the ordering induced by the utility function are broken randomly. Note that random performance according to this error measure is a score of 0.5, and that a score of 0.0 indicates a perfect ordering. All results that follow are averaged over the 45 users that provided the largest number of movie ratings. To normalize for different numbers of ratings, for each user we consider exactly 500 movie ratings randomly selected from her rating list.

The left-hand panel of Figure 1 shows how well the utility function orders the movies depending on the number of preference statements used to generate this function. In this analysis we use the top $k$ preference statements as returned by `c45rules`. Each curve in Figure 1 gives the performance for a different choice of kernel degree, i.e., different choice of kernel parameters $\lambda_1, ..., \lambda_n$. The "degree" $d$ indicates that all $\lambda_i$ with $i > d$ are set to zero, while all others are one. This eliminates all monomials of size greater then $d$. For small numbers of preference statements, all degrees perform roughly equivalently, but for larger sets of preference statements, high-degree kernels substantially outperform low-degree kernels. It appears that low-degree kernels cannot capture the dependencies in the preference statements used in the evaluation, and thus the ability to handle large-degree monomials (i.e., non-linear interactions between attributes $\mathbf{X}$) is beneficial.

Since we are using a very coarse description of the movies, the attributes do not suffice to produce a perfect ordering from a small number of preference statements. In particular, the average error rate of the complete set of C4.5 rules is 0.24. Note that this pairwise classification performed by C4.5 is potentially easier than the utility revelation problem, since the rules do not have to form an ordering. Comparing the C4.5 performance against the error rates of around 0.27 achieved by the ordinal utility function for the high-degree kernels, we conclude that our method performs the translation into a consistent ordering effectively and with good accuracy.

---

[5] http://research.compaq.com/SRC/eachmovie/
[6] http://www.cse.unsw.edu.au/∼quinlan/

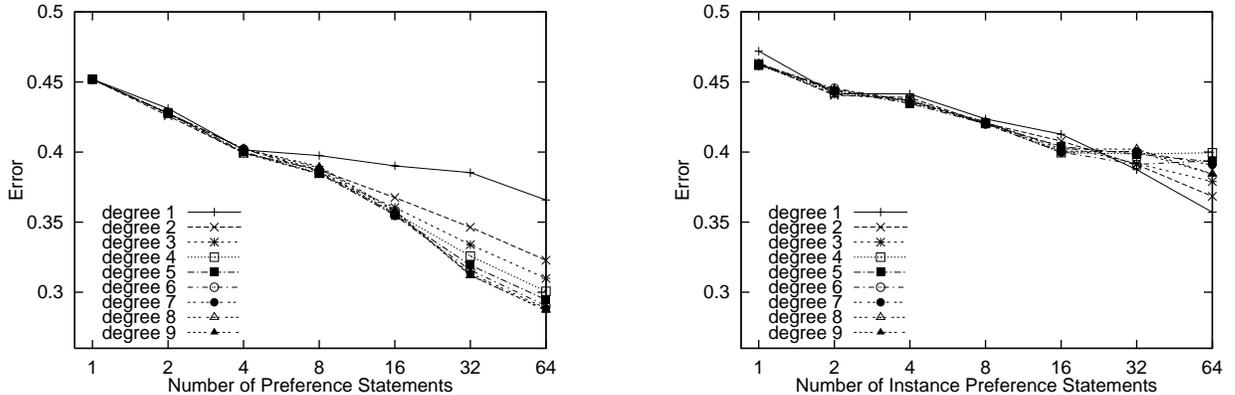

Figure 1: Average error rate as a function of the number of statements in $S$, where $S$ contains unrestricted generalizing (left), or (right) only instance level statements.

The right-hand panel shows an analog plot for using only instance-level statements. Compared to the generalizing statements, the error rates here are worse, and this indicates the beneficial expressive power of generalizing statements. For both instance and generalizing statements we performed additional experiments using a soft-margin approach. This reduced error rates, but gave qualitatively similar results. Regarding computational efficiency, the average CPU-time of $SVM^{light}$ for solving the quadratic program for a set of 64 generalizing statements was less than 0.1 seconds.

## 5 RELATED WORK AND CONCLUSIONS

We have described a novel approach to ordinal utility revelation from a set of qualitative preference statements. To the best of our knowledge, our proposal constitutes the first solution to this problem that can handle heterogeneous preference statements both efficiently and effectively. The key technical contribution is a computationally tractable, non-parametric transformation into a space where ordinal utility functions decompose linearly and where dimensions have clear and intuitive semantics. As such, our approach addresses a long-standing open question in the area of preference representation, formulated by Doyle [8] as: "Can one recast the underlying set [of attributed alternatives] in terms of a different [from the original attribution] span of dimensions such that the utility function becomes linear? If so, can one find new linearizing dimensions that also mean something to human interpreters?"

We have found in the literature only one work directly attempting to shed some light on this question, namely the work of Shoham on utility distributions [21]. Specifically, Shoham shows that a set of linearizing dimensions exists for any utility function, and that this set of dimensions may have to be exponentially larger than the original set of attributes. The result of Shoham, however, is more foundational than operational. First, the connection between the original attribution and the particular set of dimensions proposed in [21] is not generally natural, and thus it is rather unclear how to perform preference elicitation with respect to this set of dimensions. Second, no efficient computational scheme for reasoning about this set of dimensions has been proposed so far. Thus, we believe that our work is the first to provide an affirmative, practically usable, answer to the question of generic existence of an intuitive linearizing space of dimensions.

Our ongoing and future work builds upon the foundations laid in this paper in several directions. First, we would like to provide some informative upper bounds on the number of preference statements that a user will have to specify before the inferred utility function approximates her preferences sufficiently well. Furthermore, we would like to study applicability and efficiency of standard active learning techniques to mixed-initiated preference elicitation in our framework. Finally, we would like to perform a deeper analysis of the semantics of our inference procedure, connecting it, for instance, with the recent axiomatic approaches for preference revelation such as [16].

This work was funded in part under NSF CAREER Award IIS-0237381.## References

[1] F. Bacchus and A. Grove. Graphical models for preference and utility. In *UAI-95*, pages 3–10, 1995.

[2] D. Bertsekas, A. Nedic, and A. Ozdaglar. *Convex Analysis and Optimization*. Athena Scientific, 2003.

[3] C. Boutilier, F. Bacchus, and R. I. Brafman. UCP-networks: A directed graphical representation of conditional utilities. In *UAI-01*, pages 56–64, 2001.

[4] C. Boutilier, R. Brafman, C. Domshlak, H. Hoos, and D. Poole. CP-nets: A tool for representing and reasoning about conditional *ceteris paribus* preference statements. *J. Artif. Intel. Res.*, 21:135–191, 2004.

[5] R. Brafman, C. Domshlak, and T. Kogan. Compact value-function representations for qualitative preferences. In *UAI-04*, 2004.

[6] R. M. Chisholm and E. Sosa. On the logic of 'Intrinsically Better'. *American Philosophical Quarterly*, 3:244–249, 1966.

[7] C. Cortes and V. Vapnik. Support–vector networks. *Machine Learning Journal*, 20:273–297, 1995.

[8] J. Doyle. Prospects for preferences. *Computational Intelligence*, 20(2):111–136, 2004.

[9] J. Doyle and R. H. Thomason. Background to qualitative decision theory. *AI Magazine*, 20(2):55–68, 1999.

[10] P. E. Green, A. M. Krieger, and Y. Wind. Thirty years of conjoint analysis: Reflections and prospects. *Interfaces*, 31(3):56–73, 2001.

[11] V. Ha and P. Haddawy. A hybrid approach to reasoning with partially elicited preference models. In *UAI-99*, pages 263–270, 1999.

[12] S. O. Hansson. *The Structure of Values and Norms*. Cambridge University Press, 2001.

[13] R. Herbrich, T. Graepel, and K. Obermayer. Large margin rank boundaries for ordinal regression. In *Advances in Large Margin Classifiers*, pages 115–132. 2000.

[14] R. L. Keeney and H. Raiffa. *Decision with Multiple Objectives*. Wiley, 1976.

[15] D. H. Krantz, R. D. Luce, P. Suppes, and A. Tversky. *Foundations of Measurement*. Academic, 1971.

[16] P. La Mura. *Foundations of Multi-Agent Systems*. PhD thesis, Graduate School of Business, Stanford, 1999.

[17] P. La Mura and Y. Shoham. Expected utility networks. In *UAI-99*, pages 367–373, 1999.

[18] M. McGeachie and J. Doyle. Utility functions for ceteris paribus preferences. *Computational Intelligence*, 20(2):158–217, 2004.

[19] D. J. Packard. A preference logic minimally complete for expected utility maximization. *Journal of Philosophical Logic*, 4(2):223–235, 1975.

[20] J. R. Quinlan. *C4.5: Programs for Machine Learning*. Morgan Kaufmann, San Mateo, CA, 1993.

[21] Y. Shoham. A symmetric view of probabilities and utilities. In *IJCAI-97*, pages 1324–1329, 1997.

[22] V. Vapnik. *Statistical Learning Theory*. Wiley, 1998.